\documentclass[pdflatex,sn-mathphys-num]{sn-jnl}

\usepackage{graphicx}%
\usepackage{multirow}%
\usepackage{amsmath,amssymb,amsfonts}%
\usepackage{amsthm}%
\usepackage[title]{appendix}%
\usepackage{xcolor}%
\usepackage{textcomp}%
\usepackage{booktabs}%
\usepackage{array}%
\usepackage{float}%
\usepackage{placeins}%

\hypersetup{hypertexnames=false}
\begin{document}

\title[FD-DB]{FD-DB: Frequency-Decoupled Dual-Branch Network for Unpaired Synthetic-to-Real Domain Translation}

\author[1,2]{\fnm{Chuanhai} \sur{Zang}}\email{tryzangfc@gmail.com}
\author[1,2]{\fnm{Jiabao} \sur{Hu}}
\author*[1,2]{\fnm{XW} \sur{Song}}\email{songxw@zju.edu.cn}

\affil[1]{\orgname{State Key Laboratory of Fluid Power and Mechatronic System, School of Mechanical Engineering, Zhejiang University}, \orgaddress{\city{Hangzhou}, \state{Zhejiang}, \postcode{310058}, \country{China}}}
\affil[2]{\orgname{Zhejiang Key Laboratory of Advanced Manufacturing Technology, School of Mechanical Engineering, Zhejiang University}, \orgaddress{\city{Hangzhou}, \state{Zhejiang}, \postcode{310058}, \country{China}}}

\abstract{Synthetic data provide low-cost, accurately annotated samples for geometry-sensitive vision tasks, but appearance and imaging differences between synthetic and real domains cause severe domain shift and degrade downstream performance. Unpaired synthetic-to-real translation can reduce this gap without paired supervision, yet existing methods often face a trade-off between photorealism and structural stability: unconstrained generation may introduce deformation or spurious textures, while overly rigid constraints limit adaptation to real-domain statistics. We propose FD-DB, a frequency-decoupled dual-branch model that separates appearance transfer into low-frequency interpretable editing and high-frequency residual compensation. The interpretable branch predicts physically meaningful editing parameters (white balance, exposure, contrast, saturation, blur, and grain) to build a stable low-frequency appearance base with strong content preservation. The free branch complements fine details through residual generation, and a gated fusion mechanism combines the two branches under explicit frequency constraints to limit low-frequency drift. We further adopt a two-stage training schedule that first stabilizes the editing branch and then releases the residual branch to improve optimization stability. Experiments on the YCB-V dataset show that FD-DB improves real-domain appearance consistency and significantly boosts downstream semantic segmentation performance while preserving geometric and semantic structures.}

\keywords{Synthetic-to-Real (Syn2Real), Unpaired Translation, Frequency Decoupling, Dual-Branch Generative Model, Domain Adaptation}

\maketitle

\section{Introduction}\label{sec:intro}

In recent years, deep learning has become the dominant paradigm for visual perception tasks such as object detection and semantic segmentation, and has gradually been extended to problems with higher requirements on geometric accuracy, including 6D pose estimation (simultaneously regressing/solving three-dimensional rotation and three-dimensional translation)~\cite{ren2015fasterrcnn,chen2018deeplabv3,xiang2018posecnn,guan2024survey}. The performance upper bound of tasks such as semantic segmentation is often constrained by data scale, scene coverage, and annotation quality. Collecting strongly supervised annotations, e.g., accurate poses and instance-level masks, is costly and time-consuming. Moreover, such datasets are typically captured in laboratory environments and differ from real deployment scenarios, making it difficult to cover multi-source variations in illumination, material properties, sensor imaging characteristics, and background complexity~\cite{guan2024survey}.

Generating synthetic data with renderers/simulators is a common way to alleviate the annotation bottleneck. Using tools such as Blender, one can freely set backgrounds, materials, lighting, and poses, and render 3D models through automated pipelines to directly output pixel-wise masks, depth maps, and 6D pose ground truth, thereby obtaining large-scale strongly supervised training samples at relatively low cost. However, models trained purely on synthetic data often suffer from substantial performance degradation on real data. The primary reason is the ``domain gap/reality gap'' between the synthetic and real domains in terms of illumination distributions, imaging noise, and post-processing, which prevents feature distributions and decision boundaries from transferring directly~\cite{tobin2017domain,bousmalis2017pixel,hoffman2018cycada}. Therefore, without sacrificing label availability, effectively aligning the appearance statistics of synthetic images to the real-domain distribution is a key problem for improving real-world generalization in downstream detection/segmentation/pose tasks~\cite{bousmalis2017pixel,hoffman2018cycada}.

Given only an unlabeled set of real images, pixel-level domain adaptation typically relies on unpaired image-to-image translation to learn an appearance mapping from the synthetic domain to the real domain, thereby improving downstream real-domain generalization while preserving synthetic labels~\cite{bousmalis2017pixel,hoffman2018cycada,safayani2025review}. However, this problem is inherently underdetermined: once the generated results exhibit deformation, hallucination, or semantic/structural shift, label inheritability is compromised and the training benefit is reduced~\cite{imbusch2022cut,guo2022structure,safayani2025review}. Consequently, for geometry-sensitive Sim2Real applications, the key is not only to improve photorealism, but also to explicitly suppress translation behaviors that would break geometric/semantic consistency during appearance alignment, so as to ensure label inheritability and training effectiveness~\cite{imbusch2022cut,safayani2025review}.

The main contributions of this paper are summarized as follows.
\begin{itemize}
\item First, a dual-branch generator framework for unpaired Sim2Real translation is proposed, where an interpretable editing branch models low-frequency imaging style, a free residual branch complements high-frequency details, and a gating mechanism enables controllable fusion.
\item Second, a frequency-domain constrained residual injection mechanism (high-pass constraint + multi-scale low-frequency anchoring) is introduced to explicitly limit low-frequency drift and to enhance controllability and robustness under strong content-preservation requirements.
\item Third, a stage-wise stable training schedule (edit stage, free stage) is designed, where stage switching is driven by stability criteria, thereby reducing training instabilities and the risks of structural/color drift caused by the underdetermined nature of the problem.
\end{itemize}

\section{Related Work}\label{sec:related}

In Syn2Real settings, unpaired translation is inherently under-constrained, i.e., it lacks sufficient constraints and inductive bias, and therefore is prone to deformation and hallucination that render synthetic labels unusable, violating the fundamental requirement that ``the translated image should still inherit the synthetic annotations''~\cite{imbusch2022cut}. Consequently, the key challenge of unpaired translation for Syn2Real is not merely to improve photorealism, but more importantly to suppress structural drift and maintain content consistency.

To this end, many studies have introduced various forms of constraints. CyCADA jointly models pixel-level alignment and feature-level alignment, and incorporates constraints such as cycle consistency and semantic consistency to mitigate risks caused by appearance-only transfer, including semantic drift and label flipping (semantic inconsistency)~\cite{hoffman2018cycada}. CycleGAN leverages bidirectional mappings and cycle-consistency to shrink the solution space in the absence of paired supervision, making training more stable and enabling practical unpaired translation~\cite{zhu2017cyclegan}. Contrastive-learning-driven unpaired translation provides a direct and relatively lightweight route to content/structure preservation. CUT, with PatchNCE as its core, reinforces the correspondence of local patches between the input and the output in feature space, achieving unpaired translation without explicit inverse mapping and reducing architectural overhead~\cite{park2020cut}. However, these methods primarily constrain semantic and structural changes in different forms, rather than fundamentally resolving the under-constrained nature of unpaired translation; as a result, content and structural variations can still be introduced.

In this paper, an image-attribute editing paradigm is adopted (Generator: Interpretable Editing Branch (Parameter-Editing Branch) G\_edit). In essence, it adjusts imaging attributes such as exposure, and does not alter image content or structure, thereby fundamentally addressing the under-constrained issue of unpaired translation and ensuring high label usability. Yang et al. observed that many cross-domain discrepancies in imaging attributes (exposure, contrast, color cast, camera response, etc.) vary slowly in space and are typically manifested as low-frequency statistical differences; FDA can substantially reduce the domain gap and improve downstream segmentation generalization by exchanging low-frequency spectral components, indicating that low-frequency alignment directly targets domain-related appearance statistics~\cite{yang2020fda}. Both Yang et al.'s findings and the ablation studies in this paper on the interpretable editing branch consistently demonstrate that modifying image attributes alone can effectively align synthetic images toward the real domain.

Nevertheless, using only the interpretable editing branch often fails to achieve the desired performance, primarily due to limited expressiveness: it still falls short of real scenes in aspects such as local specular intensity. Studies on multi-scale/frequency decomposition provide a viable direction. Multi-scale decomposition/reconstruction has been shown effective for separating global attributes from detail refinement in image translation. For instance, Liang et al. employ a Laplacian pyramid to disentangle global structure and details, balancing efficiency and realism, and achieve strong results on high-resolution photo translation~\cite{liang2021lptn}. Cai et al. further propose a translation framework that performs low-/high-frequency decoupling directly in the frequency domain with joint constraints, showing that high-frequency components are strongly correlated with identity/structure preservation and that explicit decomposition helps reduce structural damage caused by excessive stylization~\cite{cai2021frequency}. Finally, from an optimization perspective, deep networks exhibit a spectral bias toward fitting low-frequency components first, making high-frequency details more likely to be missing or replaced by spurious textures; introducing frequency-domain losses (e.g., Focal Frequency Loss) to explicitly reduce spectral discrepancies can improve reconstruction/synthesis quality, providing additional motivation for imposing explicit constraints on high-frequency or spectral differences~\cite{rahaman2019spectral,jiang2021ffl}.

Therefore, a dual-branch generator framework for unpaired Sim2Real translation is proposed, where an interpretable editing branch models low-frequency imaging style and a free residual branch compensates high-frequency details. Furthermore, a frequency-domain constrained residual injection mechanism (high-pass constraint + multi-scale low-frequency anchoring) is introduced to explicitly restrict low-frequency drift and to enhance controllability and robustness under strong content-preservation requirements.

\section{Method}\label{sec:method}

\subsection{Problem Formulation and Overview}\label{sec:method-overview}

Let the synthetic domain and the real domain be denoted by $\mathcal{S}$ and $\mathcal{R}$, respectively. During training, only unpaired image sets are provided, i.e., real-domain images are unlabeled and no paired correspondences are available.

The objective is to learn a mapping $G:\mathcal{S}\rightarrow \mathcal{R}$ such that, for any input synthetic image $x_s$, the translated output $y_R=G(x_s)$ matches the real-domain appearance distribution $p_R$ while preserving the content structure consistent with $x_s$ (e.g., object boundaries, local geometric contours, and relative spatial layout). This setting targets geometry-sensitive downstream vision tasks; therefore, ``structural stability'' and ``appearance alignment'' must be satisfied simultaneously.

\begin{figure}[H]
\centering
\includegraphics[width=\linewidth]{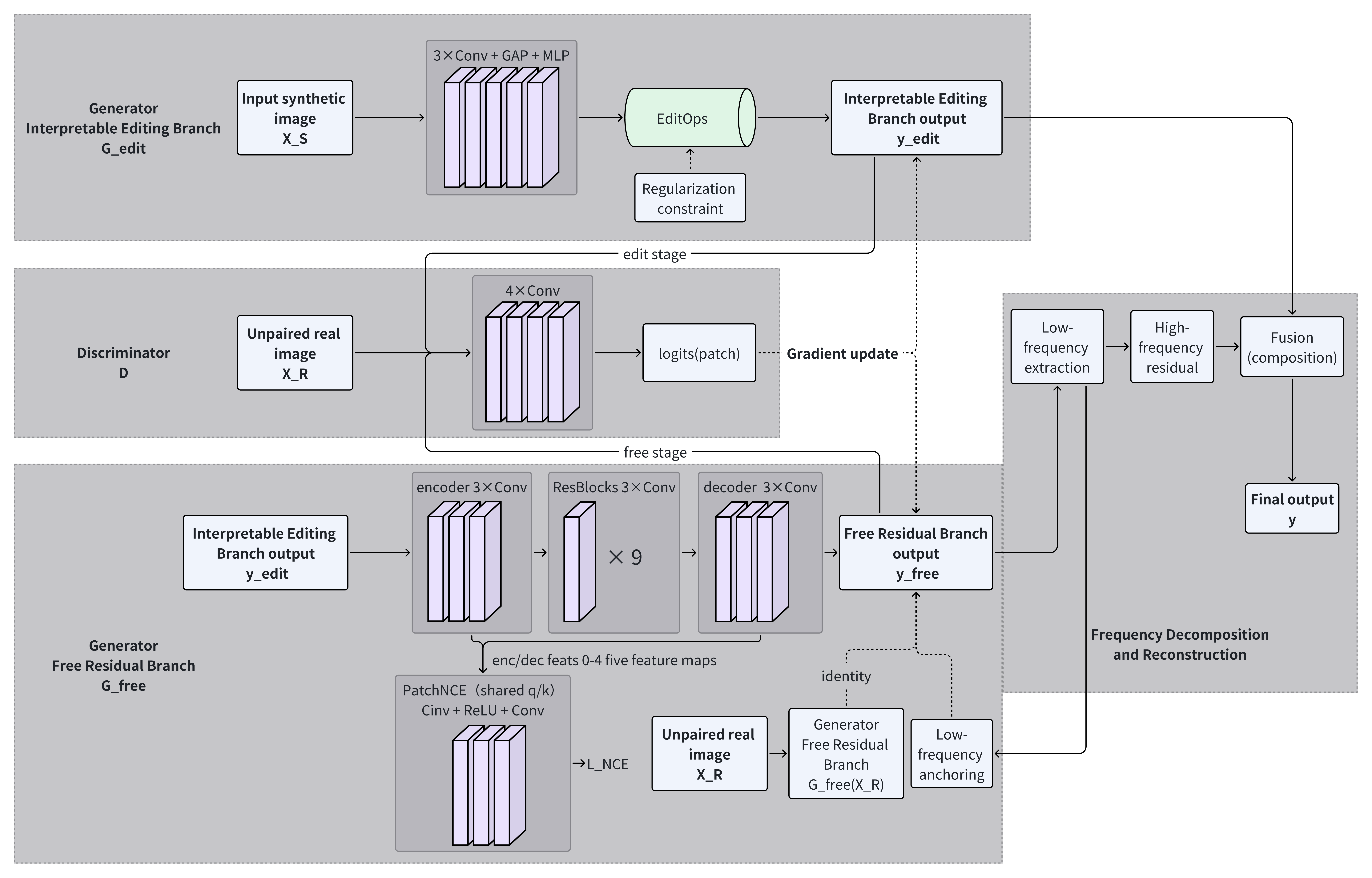}
\caption{Model architecture.}
\label{fig:arch}
\end{figure}

A frequency-decoupled dual-branch generator architecture is adopted, consisting of Generator: Interpretable Editing Branch (Parameter-Editing Branch) G\_edit and Generator: Free Residual Branch G\_free (Fig.~\ref{fig:arch}). The output of the free residual branch is processed by low-frequency extraction, and the corresponding high-frequency residual is obtained by subtraction and then added to the editing-branch output $y_{\text{edit}}$ to form the final output $y$. This part is described in detail in Sec.~\ref{sec:method-frequency}.

Within the generator, the editing branch predicts physically interpretable imaging parameters and produces a low-frequency ``base'' output of color and imaging style via a chain of differentiable editing operators (white balance, exposure, contrast, saturation, and blur/grain, etc.). The free residual branch further complements fine details and is fused with the base through a gating coefficient to obtain the final result. The motivation is to confine ``imaging attributes/color style'' primarily to the low-frequency channel, while delegating ``details'' mainly to high-frequency compensation. The two branches are detailed in Sec.~\ref{sec:method-edit} and Sec.~\ref{sec:method-free}, respectively.

The generator output and the Frequency Decomposition and Reconstruction process can be summarized as Eq.~\eqref{eq:overview_pipeline}:
\begin{equation}
\begin{aligned}
 y_{\mathrm{edit}} &= G_{\mathrm{edit}}(x_s),\\
 y_{\mathrm{free}} &= G_{\mathrm{free}}(x_s),\\
 y_{\mathrm{H}} &= y_{\mathrm{free}}-\operatorname{LP}(y_{\mathrm{free}}),\\
 y_R &= \operatorname{clip}\!\left(y_{\mathrm{edit}}+g\odot y_{\mathrm{H}}\right).
\end{aligned}
\label{eq:overview_pipeline}
\end{equation}
where input synthetic image $x_s$ is the synthetic input image; $G_{\mathrm{edit}}(\cdot)$ and $G_{\mathrm{free}}(\cdot)$ denote the two generator branches; $\operatorname{LP}(\cdot)$ is a low-pass operator; $g$ is a gate controlling the two-stage training; $y_{\mathrm{H}}$ is the high-frequency residual; and $y_R$ is the final translated image.

During training, adversarial learning is employed to align domain distributions. Discriminator D learns real-domain appearance statistics by distinguishing unpaired real image $x_R$ from generated samples $y_{\mathrm{edit}}$ and $y_{\mathrm{free}}$, thereby providing optimization signals that guide the generator to produce outputs closer to the real domain. Discriminator-related details are presented in Sec.~\ref{sec:method-disc}.

In terms of training strategy, a two-stage schedule is adopted: the gate is first fixed to suppress the free residual branch, and only the interpretable editing branch is trained to stabilize the low-frequency base; after parameters and losses satisfy a sliding-window stability criterion, the free residual branch is released to supplement high-frequency details, alleviating gradient conflicts in adversarial training and improving convergence stability. During inference, only a one-way forward pass is required to generate large batches of synthetic-to-real translated images for downstream training and evaluation. Details are provided in Sec.~\ref{sec:method-train}.

\subsection{Frequency Decomposition and Reconstruction}\label{sec:method-frequency}

\begin{figure}[H]
\centering
\includegraphics[width=\linewidth]{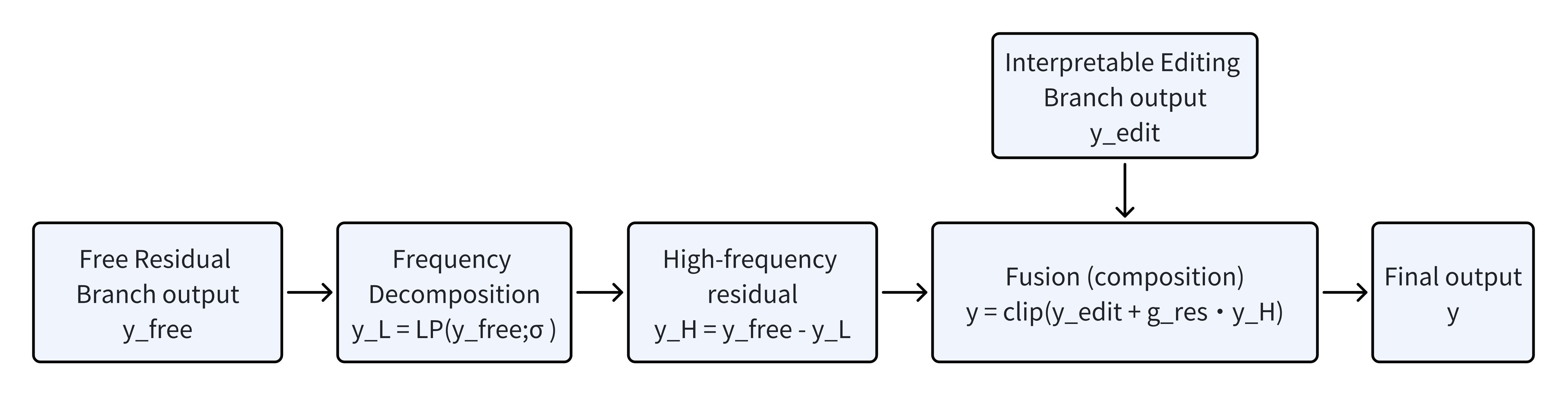}
\caption{Architecture of frequency decomposition and reconstruction.}
\label{fig:freq}
\end{figure}

To alleviate the common conflict between ``improved photorealism'' and ``structural/content stability'' in unpaired Syn2Real translation for geometry-sensitive tasks (e.g., detection, segmentation, and 6D pose estimation), an explicit decoupling is performed from a frequency-domain perspective between the ``appearance statistics that should be changed'' and the ``structural information that must be preserved'' (Fig.~\ref{fig:freq}). Specifically, low-frequency components are responsible for aligning global imaging attributes, while high-frequency components account for structural details and complex local compensation; during reconstruction, the low-frequency component is further regularized by a low-frequency anchoring constraint.

A generation mechanism consistent with the low-/high-frequency division of labor is implemented via an ``interpretable editing branch + free residual branch''. The editing-branch output $y_{\mathrm{edit}}$ serves as the base and carries the major low-frequency imaging-attribute alignment (global tone, brightness, contrast, and slowly varying appearance statistics), while the high-frequency residual extracted from the free-branch output $y_{\mathrm{free}}$ is used for high-frequency compensation. A Gaussian low-pass operator with reflection padding, $LP(\cdot;\sigma)$, is employed to perform low-frequency extraction and obtain the low-frequency component $y_L$ from $y_{\mathrm{free}}$. Compared with a hard spectral cutoff, Gaussian low-pass filtering provides better numerical stability and avoids ringing artifacts introduced by sharp cutoffs. The high-frequency residual $y_{\mathrm{H}}$ is then obtained by subtraction. The final output is obtained by fusion (composition) in Eq.~\eqref{eq:freq_reconstruct}:
\begin{equation}
 y=\operatorname{clip}\!\left(y_{\mathrm{edit}}+g_{\mathrm{res}}\cdot\bigl(y_{\mathrm{free}}-LP(y_{\mathrm{free}};\sigma)\bigr),-1,1\right),
\label{eq:freq_reconstruct}
\end{equation}
where $\operatorname{clip}(\cdot,-1,1)$ is a clipping operator that constrains each pixel value to $[-1,1]$, and $g_{\mathrm{res}}$ is used to stage-wise allocate optimization pressure across the two training phases.

\subsection{Generator: Interpretable Editing Branch (Parameter-Editing Branch) G\_edit}\label{sec:method-edit}

\begin{figure}[H]
\centering
\includegraphics[width=\linewidth]{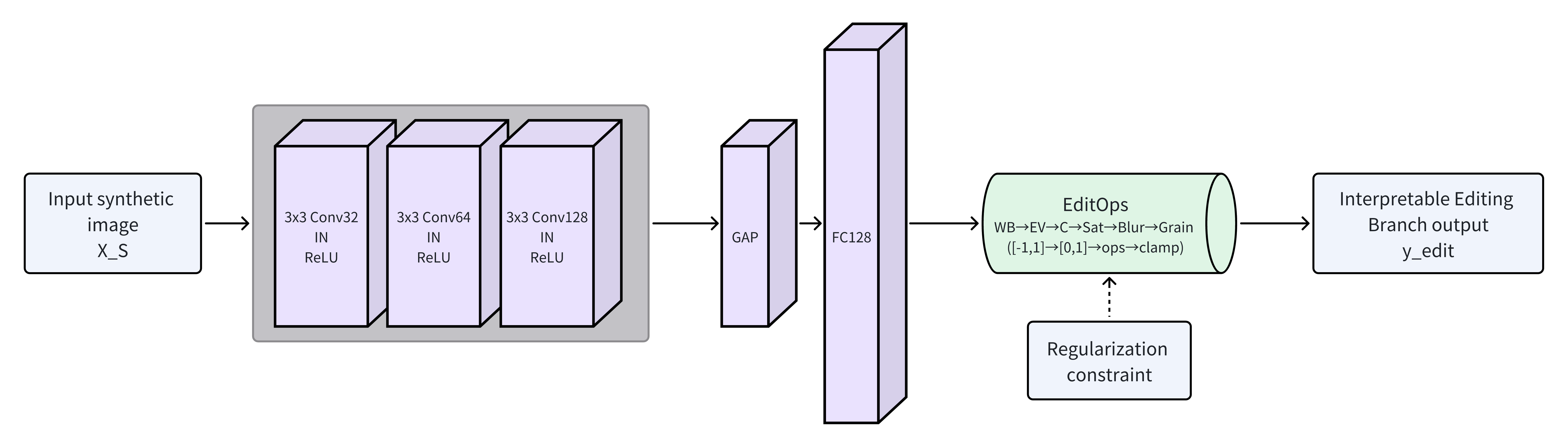}
\caption{Generator: Interpretable Editing Branch (Parameter-Editing Branch) $G_{\text{edit}}$.}
\label{fig:edit}
\end{figure}

The Parameter-Editing Branch aims to provide an interpretable, controllable, and structurally stable appearance base for the overall frequency-decoupling framework (Fig.~\ref{fig:edit}). Given input synthetic image $x_s\in[-1,1]^{C\times H\times W}$, this branch outputs editing-branch output $y_{\mathrm{edit}}$ after a parameterized imaging-operator chain, which is used to preferentially align global appearance statistics of the real domain (e.g., white balance, exposure, contrast, and saturation) while avoiding the structural drift that may arise from unconstrained generation.

This branch consists of a parameter prediction network and EditOps (Differentiable Editing Operators). The parameter prediction network regresses global editing parameters using a lightweight convolutional design: 3$\times$Conv + GAP + MLP, i.e., three $3\times 3$ convolution layers (stride$=2$) with channels $32\rightarrow64\rightarrow128$, each followed by InstanceNorm and ReLU, then a global average pooling and a two-layer MLP (hidden dimension 128) to output a raw parameter vector $r$. Each raw entry is mapped, element-wise according to a parameter specification set, into a physically interpretable parameter set $\theta$. The mapping functions include sigmoid, tanh, and log-tanh (a symmetric mapping in the log domain). Among them, log-tanh is used for positive multiplicative parameters (e.g., white-balance gains) to achieve a more reasonable scale coverage.

The editing operator chain is applied in the $[0,1]$ domain and remains end-to-end differentiable. The operators are defined as follows (Eqs.~\eqref{eq:wb}--\eqref{eq:grain}; $\widetilde{x}$ denotes a pixel value and per-pixel indices are omitted):
\begin{enumerate}
\item \textbf{White balance} (channel-wise gain $g$, $\odot$ denotes element-wise multiplication):
\begin{equation}
\widetilde{x}\leftarrow \widetilde{x}\odot g,\quad \text{map}=\text{log-tanh}.
\label{eq:wb}
\end{equation}
\item \textbf{Exposure} (EV is log exposure):
\begin{equation}
\widetilde{x}\leftarrow \widetilde{x}\cdot 2^{\mathrm{EV}},\quad \text{map}=\text{tanh}.
\label{eq:exp}
\end{equation}
\item \textbf{Contrast} ($c$ is the contrast factor around 0.5):
\begin{equation}
\widetilde{x}\leftarrow(\widetilde{x}-0.5)\cdot c+0.5,\quad \text{map}=\text{sigmoid}.
\label{eq:contrast}
\end{equation}
\item \textbf{Saturation} ($s>0$ is saturation, $w$ is RGB-to-luminance weight):
\begin{equation}
\ell=w^\top\widetilde{x},\quad \widetilde{x}\leftarrow \ell+s(\widetilde{x}-\ell),\quad \text{map}=\text{sigmoid}.
\label{eq:saturation}
\end{equation}
\item \textbf{Blur} (separable Gaussian). For each sample, a continuous value $\sigma_b$ is predicted (via sigmoid). The kernel size $K$ is determined by a preset maximum standard deviation $\sigma_{\max}$:
\begin{equation}
K=2\left\lceil3\sigma_{\max}\right\rceil+1.
\label{eq:blur_kernel}
\end{equation}
\item \textbf{Grain} (smoothed noise injection). Sample $\epsilon\sim\mathcal{N}(0,1)$, smooth it with Gaussian filtering to control spatial correlation, and add:
\begin{equation}
\widetilde{x}\leftarrow \widetilde{x}+\alpha\cdot LP(\epsilon;\sigma_g),
\label{eq:grain}
\end{equation}
where both amplitude $\alpha$ and scale $\sigma_g$ are mapped by sigmoid.
\end{enumerate}

Consistent with the overall low-frequency/high-frequency separation principle, the parameter-editing branch mainly serves the construction of a low-frequency appearance base. However, grain is not strictly a low-frequency component; in the frequency domain, it usually appears as broadband, relatively high-frequency random micro-texture. Grain is an intentionally introduced exception in this branch, motivated by implementation and empirical observations: relying only on white balance/exposure/contrast/saturation often yields outputs that look overly clean and synthetic, whereas real imaging pipelines commonly include fine-grained perturbations such as sensor noise, compression artifacts, and film grain. Incorporating grain into this branch as an amplitude-controlled, differentiable, additive perturbation can substantially improve the realism of $y_{\mathrm{edit}}$ without introducing spatial deformation or geometric rearrangement, making it more suitable for geometry-sensitive downstream tasks (segmentation and 6D pose estimation). Meanwhile, to prevent excessive parameter drift, in addition to adversarial constraints, a Regularization constraint is applied around configuration-defined reference points during the edit stage.

\subsection{Generator: Free Residual Branch G\_free}\label{sec:method-free}

\begin{figure}[H]
\centering
\includegraphics[width=\linewidth]{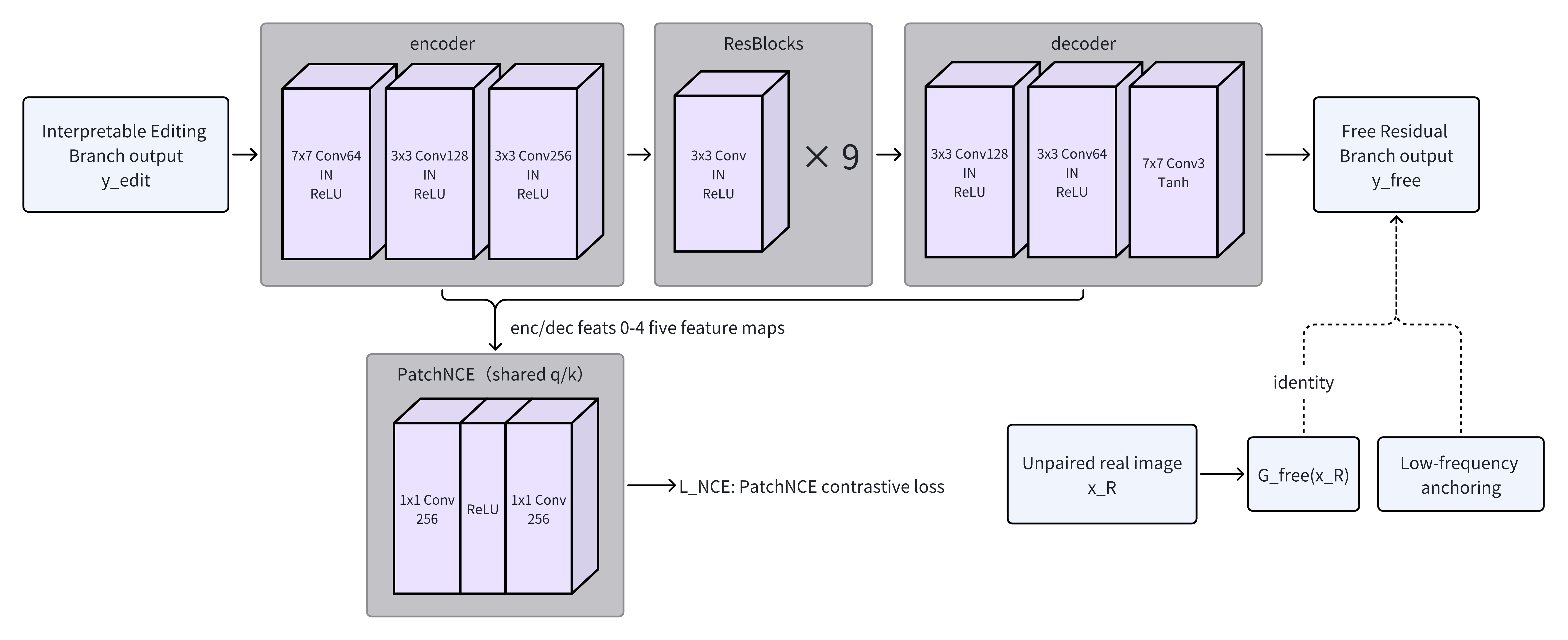}
\caption{Generator: Free Residual Branch $G_{\text{free}}$.}
\label{fig:free}
\end{figure}

Within the dual-branch frequency-decoupled framework, parameterized editing alone is insufficient to compensate for more complex local color variations, material details, and high-frequency noise statistics in the real domain, such as overexposure and specular reflections from different materials. To this end, Generator: Free Residual Branch G\_free is introduced (Fig.~\ref{fig:free}). The requirements for this branch can be summarized as follows: when the input is a synthetic image, it should modify the style to be more real-like while preserving structure and content; when the input is a real image, it should minimize unnecessary changes. This constrained free branch learns residual-style detail compensation with a more expressive generator, thereby improving realism and domain alignment with minimal damage to geometric and semantic structures. Its core architecture and training paradigm largely follow constrained Syn2Real unpaired translation~\cite{imbusch2022cut}, while adding identity and low-frequency anchoring constraints, and operating in synergy with the frequency division and two-stage scheduling proposed in this paper.

The free generator follows a CUT-style encoder 3$\times$Conv -- ResBlocks 3$\times$Conv $\times 9$ -- decoder 3$\times$Conv pipeline and outputs free-branch output $y_{\text{free}}$. Discriminator D adopts PatchGAN to emphasize local texture discrimination, thereby imposing adversarial constraints on real-domain fine-detail statistics. The training objective consists of two core constraints. The first is adversarial loss, which drives $y_{\text{free}}$ to approach the real-domain distribution from the discriminator's perspective. The second is content-preservation loss, i.e., L\_NCE via PatchNCE (shared q/k) Conv + ReLU + Conv over enc/dec feats 0--4 (five feature maps), aligning local patch representations between input and output to suppress geometric and semantic drift. Overall, this paradigm inherits the reference idea of constraining free generation via contrastive learning, enabling more stable content preservation and texture transfer under the unpaired setting.

It is worth noting that, although the free branch increases expressiveness, it can still introduce substantial appearance shifts in geometry-sensitive Syn2Real tasks, particularly global color/brightness drift and unnecessary style overfitting. To further restrict the degrees of freedom and clarify branch responsibilities, two additional constraints are introduced (Eqs.~\eqref{eq:llow} and \eqref{eq:lid}):
\begin{enumerate}
\item \textbf{low-frequency anchoring} $\mathcal{L}_{\text{low}}$. Low-frequency components at different bandwidths are extracted using multi-scale low-pass operators $LP(\cdot;\sigma_i)$, and the low-frequency content of final output $y$ is aligned with editing-branch output $y_{\text{edit}}$:
\begin{equation}
\mathcal{L}_{\text{low}}=\sum_i w_i\,\left\lVert LP(y;\sigma_i)-LP(y_{\text{edit}};\sigma_i)\right\rVert_1.
\label{eq:llow}
\end{equation}
\item \textbf{identity} $\mathcal{L}_{\text{id}}$. An unpaired real image $x_R$ is fed through the same free branch: $G(x_R)$ pathway to obtain $\hat{x}_R$, and unnecessary changes are penalized with a conservative pixel-domain term:
\begin{equation}
\mathcal{L}_{\text{id}}=\left\lVert x_R-\hat{x}_R\right\rVert_1,\quad \hat{x}_R=\text{Free branch: }G(x_R).
\label{eq:lid}
\end{equation}
\end{enumerate}
Together, these constraints explicitly restrict the free branch to a constrained generation regime that primarily aligns high-frequency details without excessively rewriting low-frequency appearance, while retaining its ability to compensate fine details.

\subsection{Discriminator D and Domain Alignment}\label{sec:method-disc}

\begin{figure}[H]
\centering
\includegraphics[width=\linewidth]{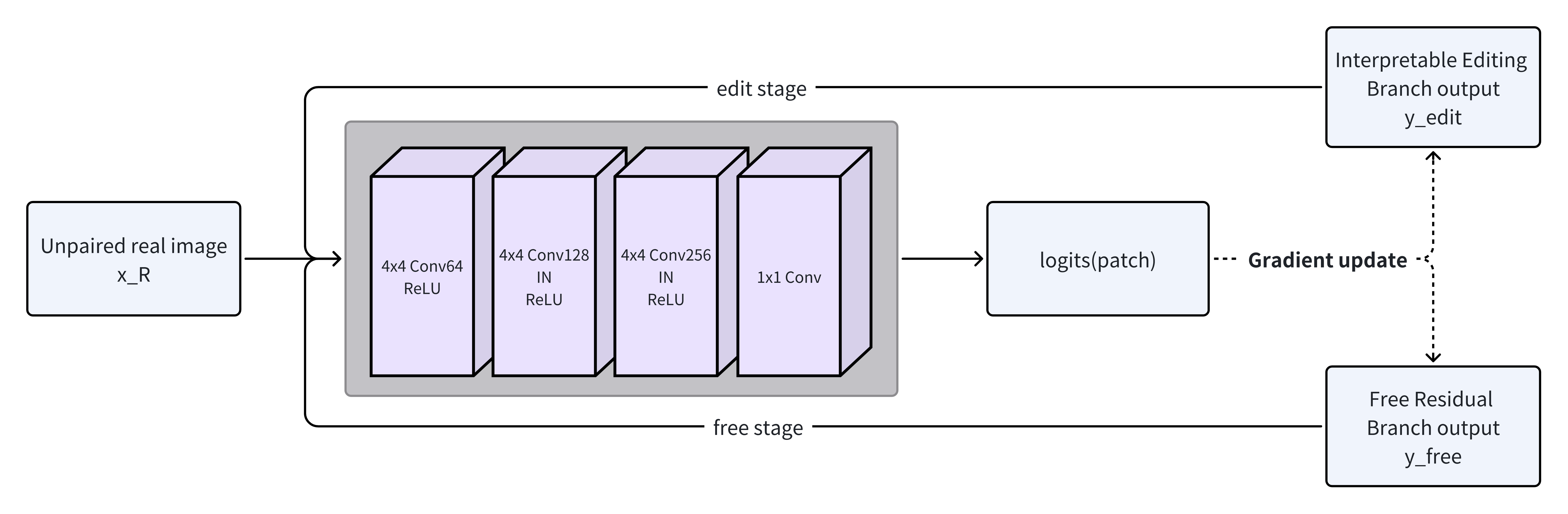}
\caption{Discriminator architecture.}
\label{fig:disc}
\end{figure}

In unpaired Syn2Real translation, the core role of Discriminator D is to provide generated samples with stable and targeted learning signals of real-domain appearance statistics (Fig.~\ref{fig:disc}). In this paper, adversarial learning is used to align distributions at the appearance level. The responsibility of D is explicitly limited to texture, noise, sharpness, and local contrast statistics, while geometric-structure preservation is enforced by the content-consistency and frequency constraints defined above, preventing adversarial signals from directly driving structural drift.

Discriminator D uses an unconditional patch-based architecture. Given a single image as input, it outputs logits(patch), where each spatial location corresponds to a realism score for one local receptive-field patch. The backbone follows 4$\times$Conv blocks with InstanceNorm and LeakyReLU (slope 0.2), and the final $1\times1$ convolution preserves logits(patch) output. This design concentrates alignment pressure on local appearance statistics and is consistent with the branch-wise division of labor.

The discriminator does not simply supervise both generation paths simultaneously. Instead, stage-wise Gradient update allocation separates responsibilities: it first aligns the controllable appearance base in the edit stage, and then, after stabilization, aligns fine-grained texture statistics in the free stage. This strategy substantially reduces gradient conflicts and role entanglement in adversarial learning.

\subsection{Optimization Objectives and Constraints}\label{sec:method-loss}

Under the unpaired Syn2Real setting, adversarial alignment is used as the appearance driver, contrastive constraints are used as the content-stability mechanism, and low-frequency anchoring with interpretable-parameter regularization is used to realize low-frequency responsibility assignment and stable training. Since training follows a two-stage schedule, the overall objective is organized in the form of base weight times stage multiplier (Eq.~\ref{eq:total_loss}):
\begin{equation}
\mathcal{L}_{G} = \lambda_{\text{gan}} m_{\text{gan}}\mathcal{L}_{\text{gan}} + \lambda_{\text{nce}} m_{\text{nce}}\mathcal{L}_{\text{nce}} + \lambda_{\text{id}} m_{\text{id}}\mathcal{L}_{\text{id}} + \lambda_{\text{edit}} m_{\text{edit}}\mathcal{L}_{\text{edit}} + \lambda_{\text{low}} m_{\text{low}}\mathcal{L}_{\text{low}},
\label{eq:total_loss}
\end{equation}
where $m_{\cdot}$ is provided by the stage controller and $\lambda_{\cdot}$ denotes global base weights.

\begin{itemize}
\item \textbf{Adversarial alignment} $\mathcal{L}_{\text{gan}}$ uses the local logits(patch) of Discriminator D for domain distribution alignment. In implementation, $D(\cdot)$ outputs logits without sigmoid, and $\mathbb{E}[\cdot]$ averages over batch and spatial positions~\cite{isola2017pix2pix,zhu2017cyclegan}. In the edit stage, a hinge adversarial loss is used~\cite{miyato2018spectral}; in the free stage, a logistic (non-saturating) adversarial loss implemented as BCEWithLogits is used, consistent with the original GAN objective~\cite{goodfellow2014gan}. The overall unpaired regime follows the GAN + PatchNCE formulation for Syn2Real translation~\cite{imbusch2022cut}.
\item \textbf{Content preservation} $\mathcal{L}_{\text{nce}}$ (denoted as \texttt{L\_NCE}) adopts PatchNCE over enc/dec feats 0--4 (five feature maps), and by default the key features stop gradients to avoid representational collapse. For each query patch feature $q$ (from generated results), the key feature $k^{+}$ at the same spatial location (from input) is treated as the positive sample, while sampled $\{k_j\}$ are negatives. The contrastive objective is given in Eq.~\eqref{eq:lnce}:
\begin{equation}
\mathcal{L}_{\text{nce}} = -\mathbb{E}\left[\log\frac{\exp(q^{\top}k^{+}/\tau)}{\sum_j \exp(q^{\top}k_j/\tau)}\right].
\label{eq:lnce}
\end{equation}
\item Identity-NCE is additionally computed on target-domain images and is enabled only in the free stage to further strengthen structural consistency.
\item \textbf{identity and low-frequency anchoring} $\mathcal{L}_{\text{id}}$ suppresses unnecessary modifications for unpaired real image $x_R$, while $\mathcal{L}_{\text{low}}$ enforces multi-scale low-pass consistency between final output $y$ and editing-branch output $y_{\text{edit}}$ in the $[0,1]$ intensity domain, explicitly assigning low-frequency to the editing branch and high-frequency to the free branch.
\item \textbf{Editing regularization} $\mathcal{L}_{\text{edit}}$ is a Regularization constraint that limits the deviation magnitude of interpretable parameters and imposes additional penalties on high-risk terms (blur and grain). For each enabled operator parameter $\theta_k$, the mean $\ell_1$ deviation from identity reference $\theta_{k,\text{id}}$ is computed, and $\mathbb{E}[|\theta_{\text{blur}}|]$, $\mathbb{E}[|\theta_{\text{grain\_amp}}|]$, and $\mathbb{E}[|\theta_{\text{grain\_size}}|]$ are added to suppress excessive degradation. This term is active only in the edit stage; it is set to zero in the free stage and the editing branch is frozen.
\end{itemize}

The above combination reflects the intended training logic: first establish a stable real-domain base in low-frequency interpretable space, then complement high-frequency and local details under joint adversarial and contrastive constraints, while continuously preventing the free branch from rewriting low-frequency statistics through low-frequency anchoring (Eq.~\ref{eq:llow}) and identity preservation (Eq.~\ref{eq:lid}). A complete notation summary is provided in Table~\ref{tab:notation} (Appendix~\ref{app:notation}).

\subsection{Training and Inference}\label{sec:method-train}

Adversarial domain-alignment training is performed under the unpaired-data setting. In each iteration, one mini-batch is sampled from the synthetic domain and one mini-batch is sampled from the real domain, maintaining an approximately 1:1 domain sampling ratio to balance gradients in adversarial training. All inputs are processed with a unified \texttt{Resize}$\rightarrow$\texttt{Normalize} preprocessing pipeline. Training is conducted on a machine equipped with an NVIDIA GeForce RTX 3060 GPU (12GB VRAM) and an AMD Ryzen 5 3600 6-core CPU. The model is implemented in PyTorch with CUDA acceleration, and supports mixed-precision training (AMP) to improve training efficiency.

A two-stage training schedule is adopted, whose core purpose is to suppress the high-degree-of-freedom updates of the free branch before the parameterized low-frequency base becomes stable, thereby reducing the direct conflict between improved photorealism and structural/content stability. The scheduler uses a sliding-window stability criterion to trigger stage switching.

After training, synthetic-domain images are migrated to the real domain through an inference pipeline to support downstream tasks. Input synthetic RGB images are first standardized and normalized into tensor representations in the $[-1,1]$ range; then, under a setting that preserves the original resolution, they are directly fed into the trained generator network for forward inference.

\section{Experiments and Evaluation}\label{sec:exp}

\subsection{Experimental Setup}\label{sec:exp-setup}

The ultimate goal of this paper is to leverage unlabeled real images to transform synthetic data, and to train downstream tasks such as semantic segmentation on the resulting dataset for direct deployment in real scenes, with performance approaching or even surpassing models trained on real datasets, as well as models trained on synthetic data and then fine-tuned on a small amount of real data. However, models for semantic segmentation typically impose stringent requirements on content invariance, and visual inspection does not necessarily correlate with practical performance. Therefore, directly benchmarking downstream task performance is essential.

To assess the effectiveness of this method, evaluation is conducted on the semantic segmentation task. YOLOv8n-seg~\cite{jocher2026yolo} is adopted as the backbone and training is conducted on the YCB-V dataset~\cite{xiang2018posecnn}, which contains 21 distinct objects with diverse shapes and is widely used in BOP-style 6D evaluation settings~\cite{hodan2018bop}. For evaluation, IoU and mIoU are used, defined in Eq.~\ref{eq:iou_miou}:
\begin{equation}
\text{IoU}_c = \frac{|P_c \cap G_c|}{|P_c \cup G_c|},\quad \text{mIoU} = \frac{1}{C}\sum_{c=1}^{C} \text{IoU}_c.
\label{eq:iou_miou}
\end{equation}

Semantic segmentation performance is evaluated under several training settings: training on real data, training on synthetic data, training on synthetic data followed by real-data fine-tuning, and training on translated data generated by this model. Moreover, since the proposed model contains two branches and the constrained free branch is developed by modifying the generator in~\cite{imbusch2022cut}, edit-only, free-only, and edit+free settings are also evaluated to compare branch-wise contributions and provide ablations.

In summary, six core settings are considered: (1) \textit{real}: training downstream tasks using only real data, serving as a practical upper bound; (2) \textit{synthetic}: training using only synthetic data, serving as a lower bound; (3) \textit{synthetic\_real}: synthetic pretraining followed by real-data fine-tuning with only 100 real samples; (4) \textit{free\_only}: translated data generated by enabling only G\_free, used to examine the impact of high-degree-of-freedom generation on geometry-sensitive tasks; (5) \textit{edit\_only}: translated data generated by enabling only G\_edit, used to validate the contribution of controllable low-frequency alignment; (6) \textit{edit\_free}: the full two-stage strategy, first converging the edit stage to stabilize low frequency and structure, then enabling the free stage to compensate real-domain high-frequency statistics.

In addition, the effect of different high-pass strengths in frequency decomposition on final results is investigated. Here, high-pass strength refers to the Gaussian kernel standard deviation $\sigma$ (in pixels). A smaller $\sigma$ yields a blur closer to the original image, so the low-pass component differs less from the input and high-frequency residual energy is smaller; a larger $\sigma$ produces stronger blur and smoother low-pass components, removing more low-frequency content and yielding higher residual energy. Six experiments are conducted with $\sigma\in\{1,2,4,8,16,32\}$, denoted as edit\_free\_hp$\sigma$.

\subsection{Results and Analysis}\label{sec:exp-results}

\begin{figure}[H]
\centering
\includegraphics[width=\linewidth]{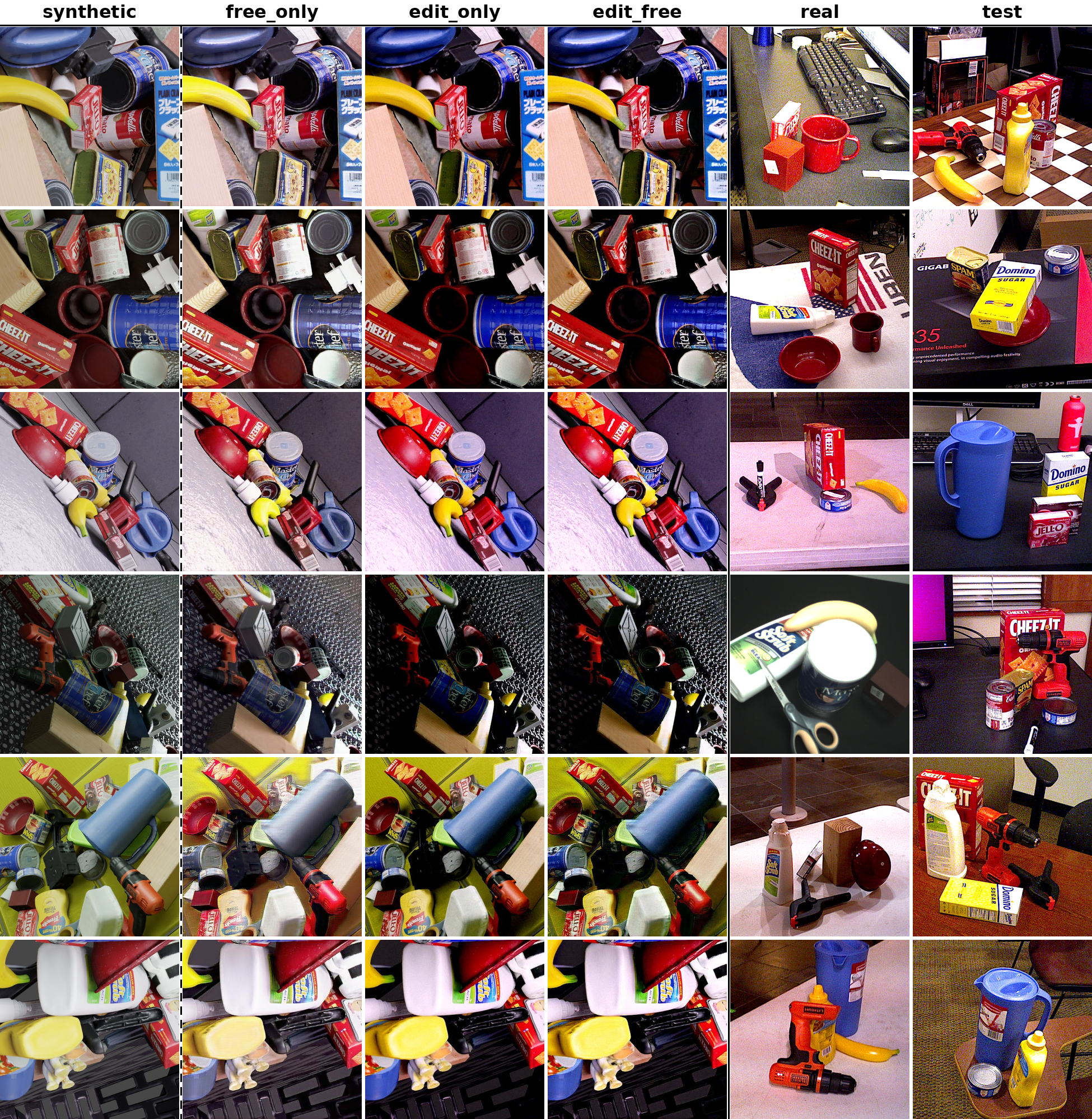}
\caption{Qualitative comparison across six experimental groups.}
\label{fig:qualitative}
\end{figure}

\begin{figure}[H]
\centering
\includegraphics[width=\linewidth]{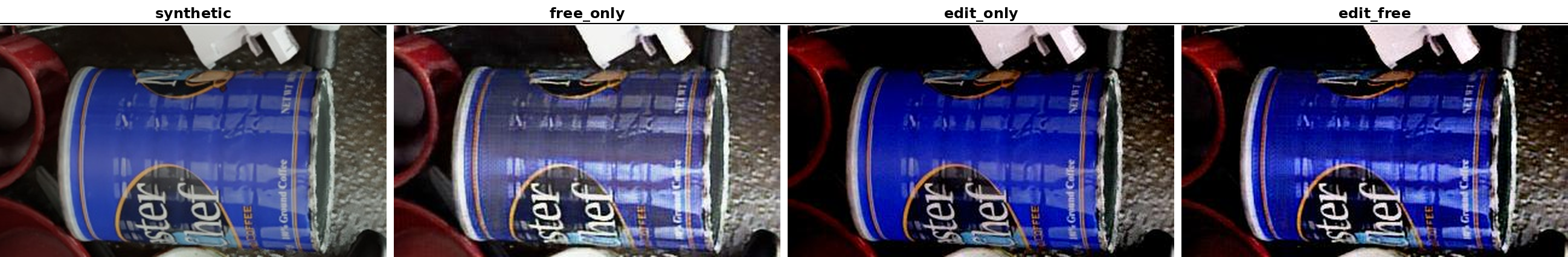}
\caption{Zoomed-in comparison of synthetic, free\_only, edit\_only and edit\_free outputs.}
\label{fig:zoom}
\end{figure}

First, for the six comparative settings, qualitative results are shown in Fig.~\ref{fig:qualitative}. From left to right, the images correspond to synthetic, free\_only, edit\_only, edit\_free (hp$=8$), real (train), and real (test). Fig.~\ref{fig:zoom} presents zoomed-in comparisons.

From the qualitative comparisons, relative to input synthetic image $x_s$, editing-branch output $y_{\text{edit}}$ aligns low-frequency attributes such as color and saturation more closely to unpaired real image $x_R$ (e.g., the blue can and kettle should exhibit noticeably higher saturation). The free branch provides higher flexibility in handling local phenomena such as specular highlights, for instance, the smooth reflective appearance on the blue can. In rare cases, this high degree of freedom can still lead to color smearing or unintended modifications (e.g., yellow-background cases with boundary smearing). Nevertheless, after high-frequency filtering, smearing is reduced. This effect becomes more evident in the residual comparisons under different high-pass strengths.

\begin{figure}[H]
\centering
\includegraphics[width=\linewidth]{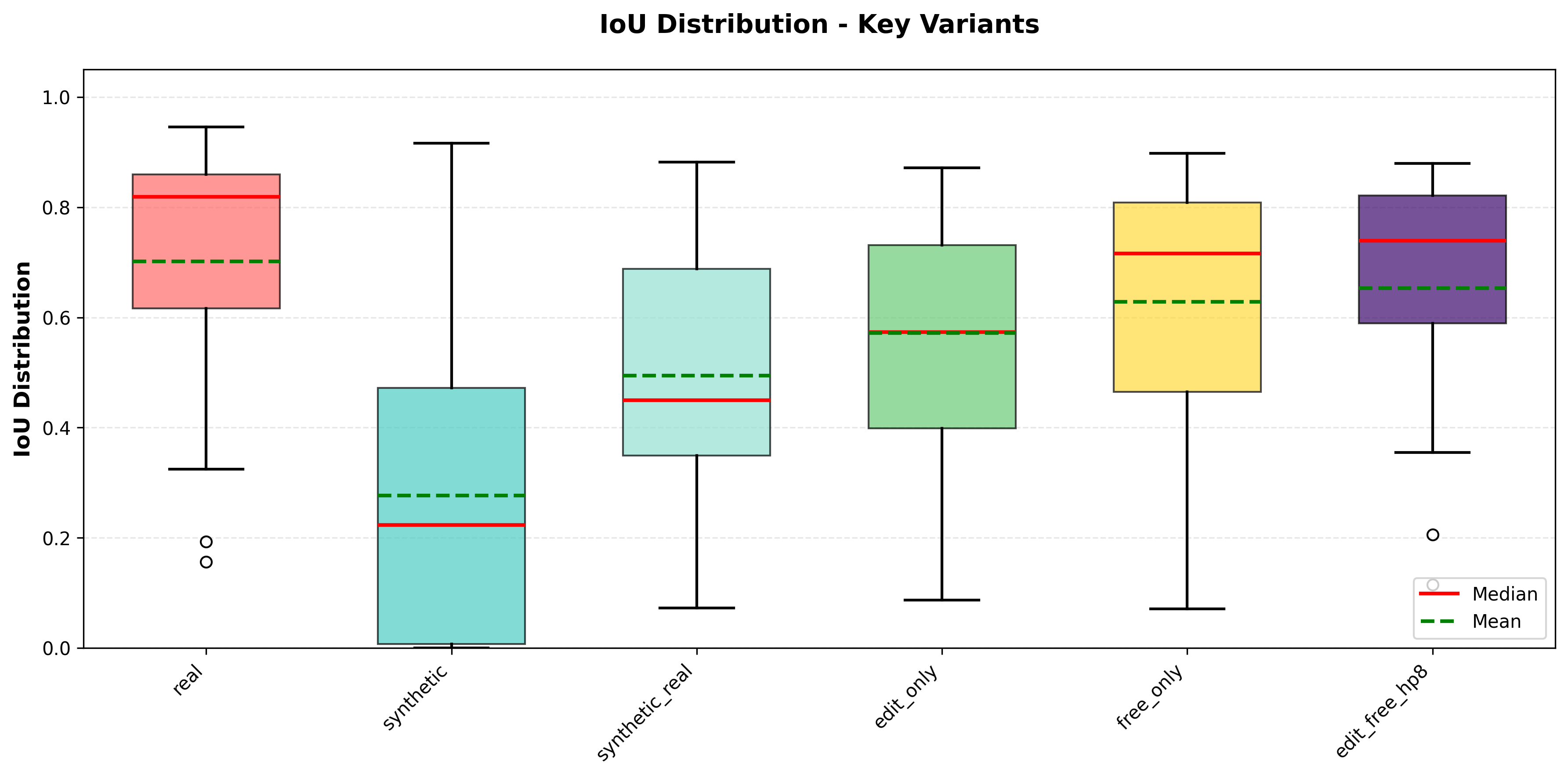}
\caption{IoU distribution boxplot for six groups (boxes show middle 50\% of classes, whiskers 1.5 IQR, dots are outliers).}
\label{fig:iou_box}
\end{figure}

Table~\ref{tab:miou} reports downstream semantic-segmentation results, and Fig.~\ref{fig:iou_box} further visualizes class-wise IoU distributions for the six-group comparison. Compared with other settings, models trained only on synthetic data have very low mIoU and poor stability across object categories, again confirming severe real-domain degradation under synthetic-only training. After pretraining on synthetic data and fine-tuning with a small amount of real data (100 images), mIoU improves but still remains far below real-only training. When larger fine-tuning sets are used, further improvement can be observed; however, such settings require substantially more real data and weaken the practical motivation of synthetic pretraining plus limited real fine-tuning.

Results from enabling each branch individually show substantial gains in both performance and stability over the synthetic baseline. Since the proposed architecture is centered on these two branches, these results both verify each branch's effectiveness and serve as mutual ablation studies. Comparing with the joint edit\_free setting further shows that frequency decomposition plus recomposition yields clear benefits. In terms of mIoU mean, median, and variance, performance is already close to real-only training.

\begin{table}[!h]
\caption{Semantic segmentation mIoU on YCB-V.}\label{tab:miou}
\centering
\begin{tabular}{lll}
\toprule
Group & Description & mIoU \\
\midrule
real & Real data only & 0.7018 \\
synthetic & Synthetic data only & 0.2768 \\
synthetic\_real & Synthetic training + real fine-tuning & 0.4942 \\
free\_only & Free branch only & 0.6283 \\
edit\_only & Editing branch only & 0.5720 \\
edit\_free\_hp8 & Two-stage edit+free (hp\,$=8$) & 0.6533 \\
\bottomrule
\end{tabular}
\end{table}
\FloatBarrier

\begin{figure}[H]
\centering
\includegraphics[width=\linewidth]{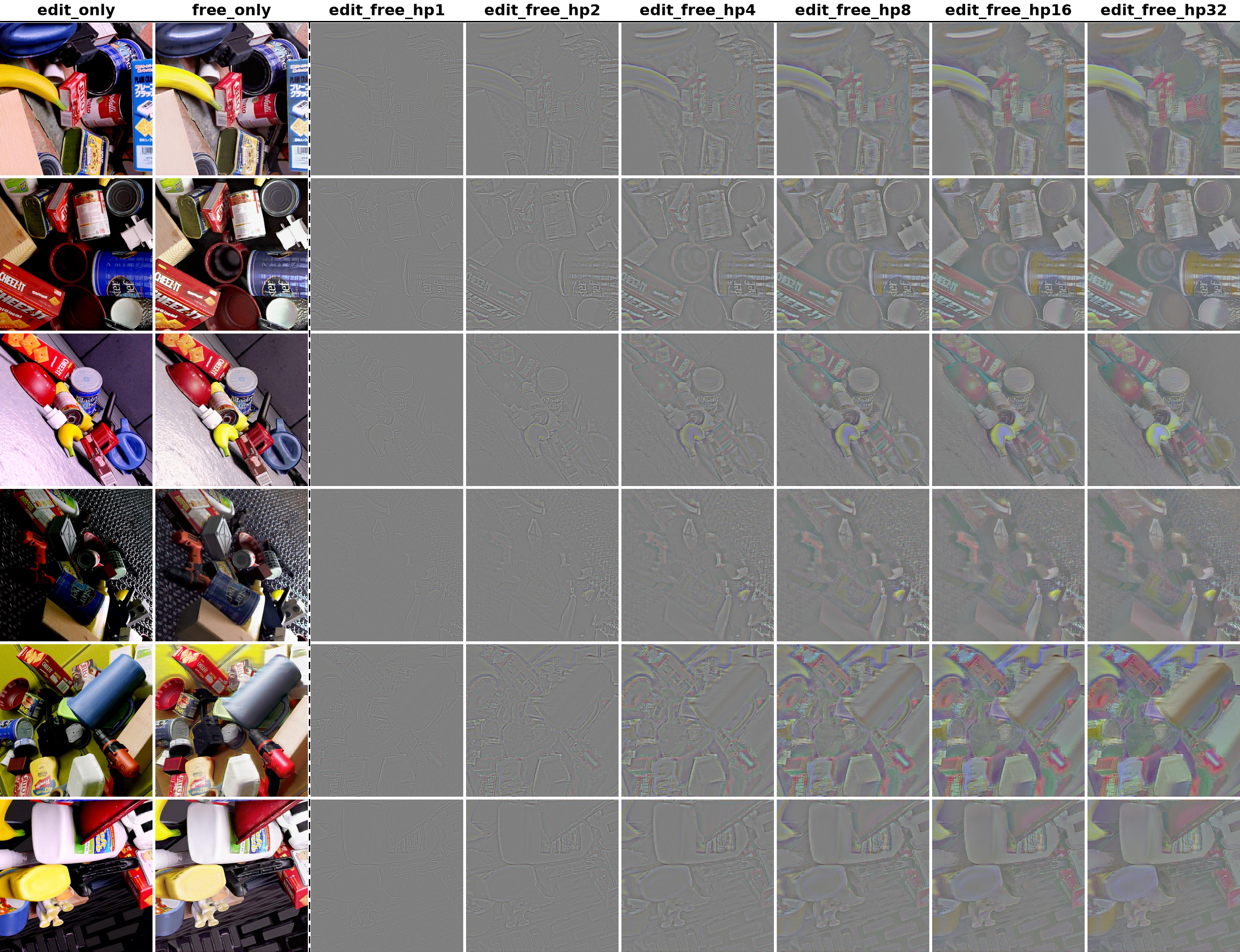}
\caption{Residual visualizations under six high-pass strengths.}
\label{fig:hp_residuals}
\end{figure}

\begin{figure}[H]
\centering
\includegraphics[width=\linewidth]{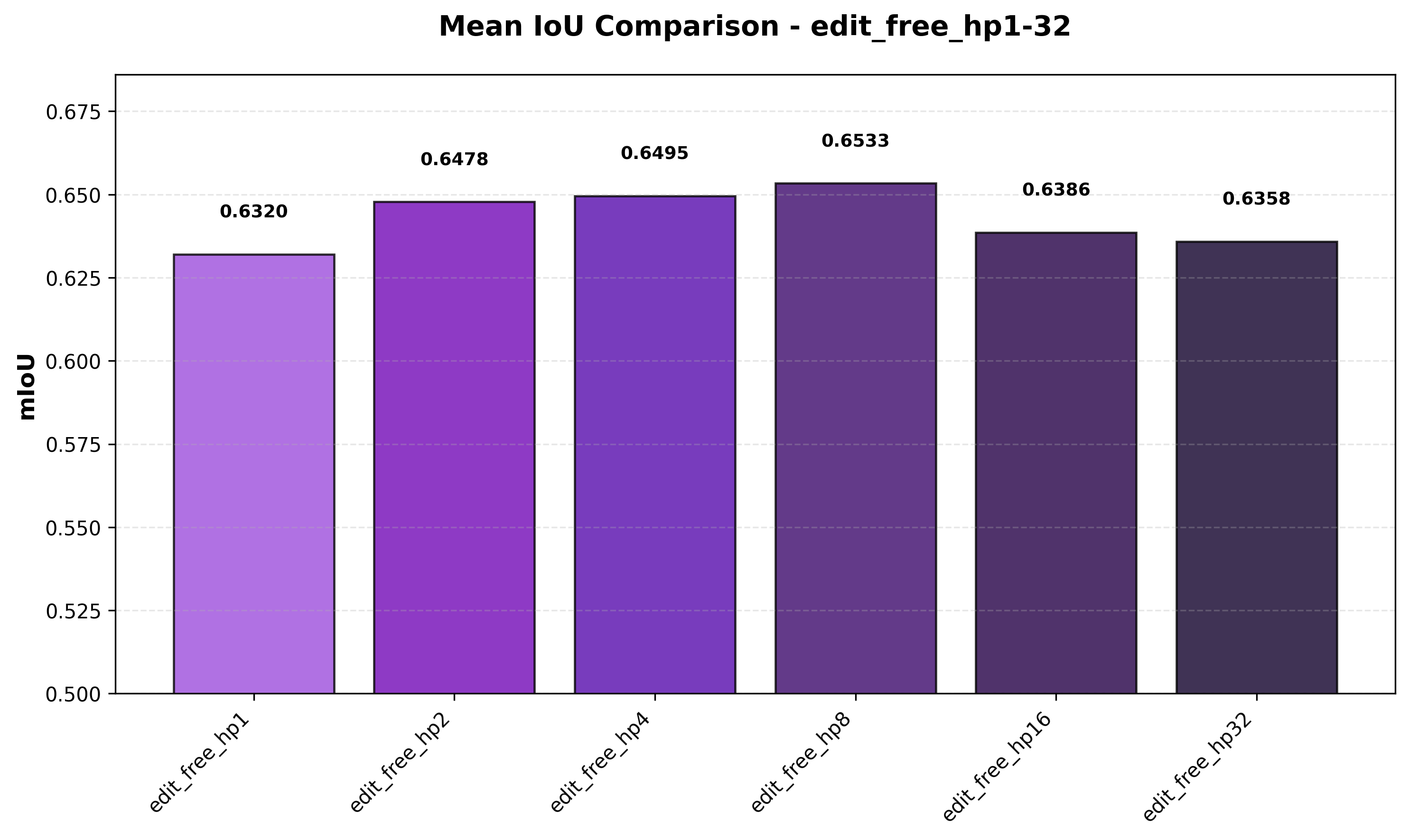}
\caption{mIoU comparison across six high-pass strengths.}
\label{fig:hp_miou}
\end{figure}

To further investigate the best high-pass strength for frequency decoupling, additional settings are evaluated. Residual visualizations and mIoU results are shown in Fig.~\ref{fig:hp_residuals} and Fig.~\ref{fig:hp_miou}. Different high-pass strengths produce substantial visual differences in residuals added to the edit base. Larger high-pass strength (larger $\sigma$) introduces more low-frequency leakage into residuals and therefore stronger large-area color shifts and smearing. Smaller high-pass strength reduces such large-area color modifications and concentrates on boundaries/local details, but too small a value may carry insufficient information. Therefore, a balance point is needed. In these evaluations, the best performance is achieved around hp$=8$.

\begin{figure}[H]
\centering
\includegraphics[width=\linewidth]{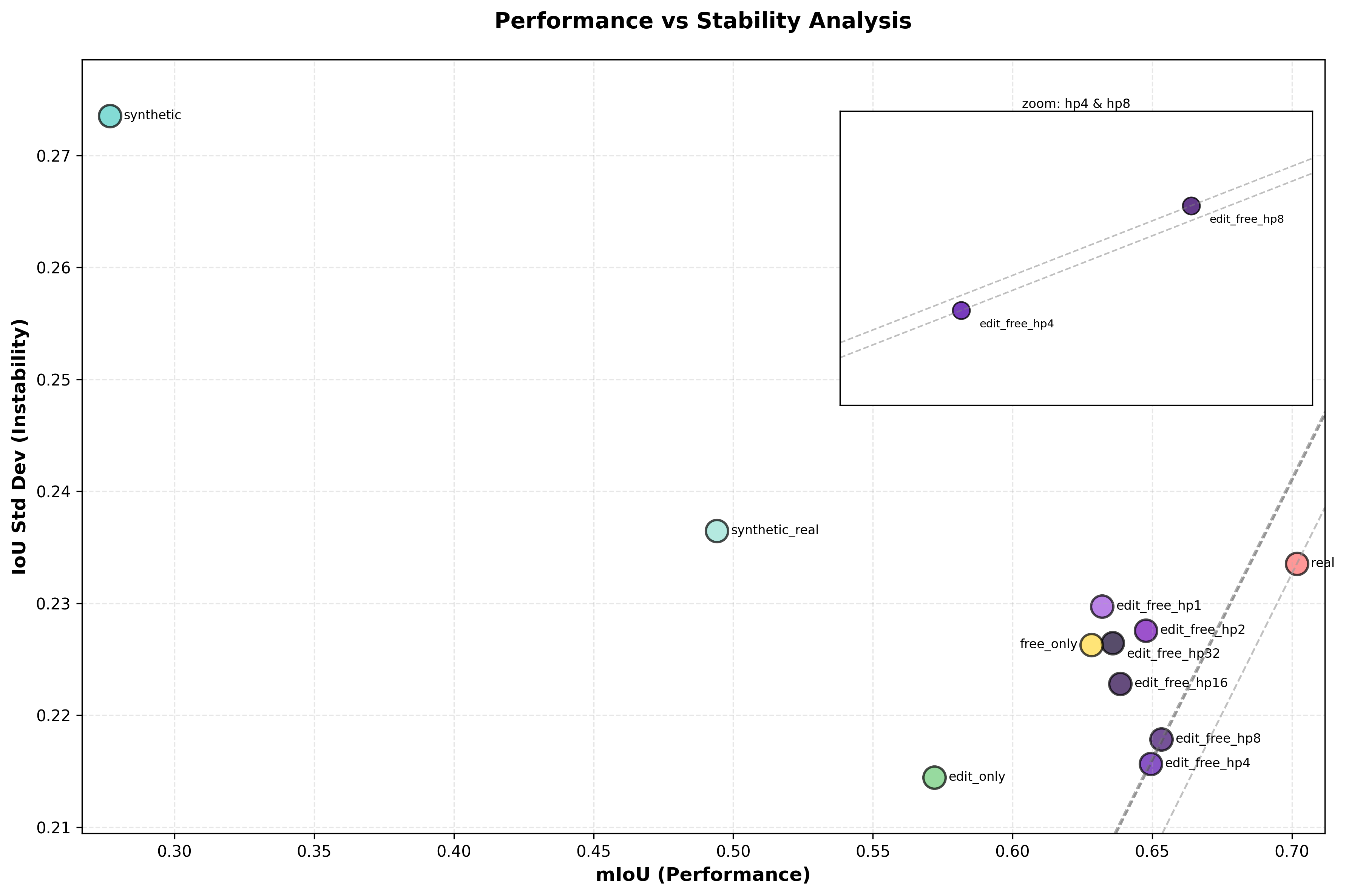}
\caption{Performance--stability trade-off plot (diagonal lines have slope $k=1$; lower-right is better).}
\label{fig:tradeoff}
\end{figure}

From the performance--stability scatter plot in Fig.~\ref{fig:tradeoff}, real-only training remains best overall when jointly considering accuracy and stability. The edit\_free variants with hp$=4$ and hp$=8$ are close, while hp$=8$ is slightly better. Additional class-level evidence is provided in Appendix~\ref{app:suppfig}, including Fig.~\ref{fig:perclass_box} and Fig.~\ref{fig:heatmap}.
\FloatBarrier

\section{Conclusion}\label{sec:conclusion}

This paper proposes a frequency-decoupled dual-branch unpaired translation model (Frequency-Decoupled Dual-Branch Unpaired Translation, FD-DB), which explicitly decomposes appearance transfer into low-frequency interpretable parameterized editing and high-frequency free residual compensation, thereby narrowing discrepancies in color/illumination and texture distributions while preserving geometric and semantic structures. Experiments on the YCB-V dataset and downstream semantic segmentation demonstrate that FD-DB improves real-domain appearance consistency and significantly boosts downstream segmentation accuracy, achieving performance close to models trained on real data and substantially outperforming training on synthetic data followed by fine-tuning with a small amount of real data.

\section*{Statements and Declarations}

\subsection*{Funding}
This research received no specific grant from any funding agency in the public, commercial, or not-for-profit sectors.

\subsection*{Competing Interests}
The authors declare that they have no competing interests.

\subsection*{Author Contributions}
Chuanhai Zang: Conceptualization, Methodology, Software, Experiments, Writing---original draft.  
Jiabao Hu: Validation, Formal analysis, Writing---review and editing.  
XW Song: Supervision, Project administration, Writing---review and editing.

\subsection*{Ethics Approval}
Not applicable.

\subsection*{Consent to Participate}
Not applicable.

\subsection*{Consent for Publication}
All authors have read and approved the final manuscript for publication.

\subsection*{Data Availability}
The YCB-Video (YCB-V) dataset used in this study is publicly available at https://bop.felk.cvut.cz/datasets/. Processed data and evaluation outputs supporting the findings of this work are available from the corresponding author upon reasonable request.

\subsection*{Code Availability}
Code and configuration files are publicly available at https://github.com/tryzang/FD-DB.

\subsection*{Use of AI Tools}
No generative AI tools were used to produce experimental results, analyses, or scientific conclusions in this manuscript.

\clearpage
\FloatBarrier
\begin{appendices}
\section{Supplementary Figures}\label{app:suppfig}

\begin{figure}[H]
\centering
\includegraphics[width=\linewidth]{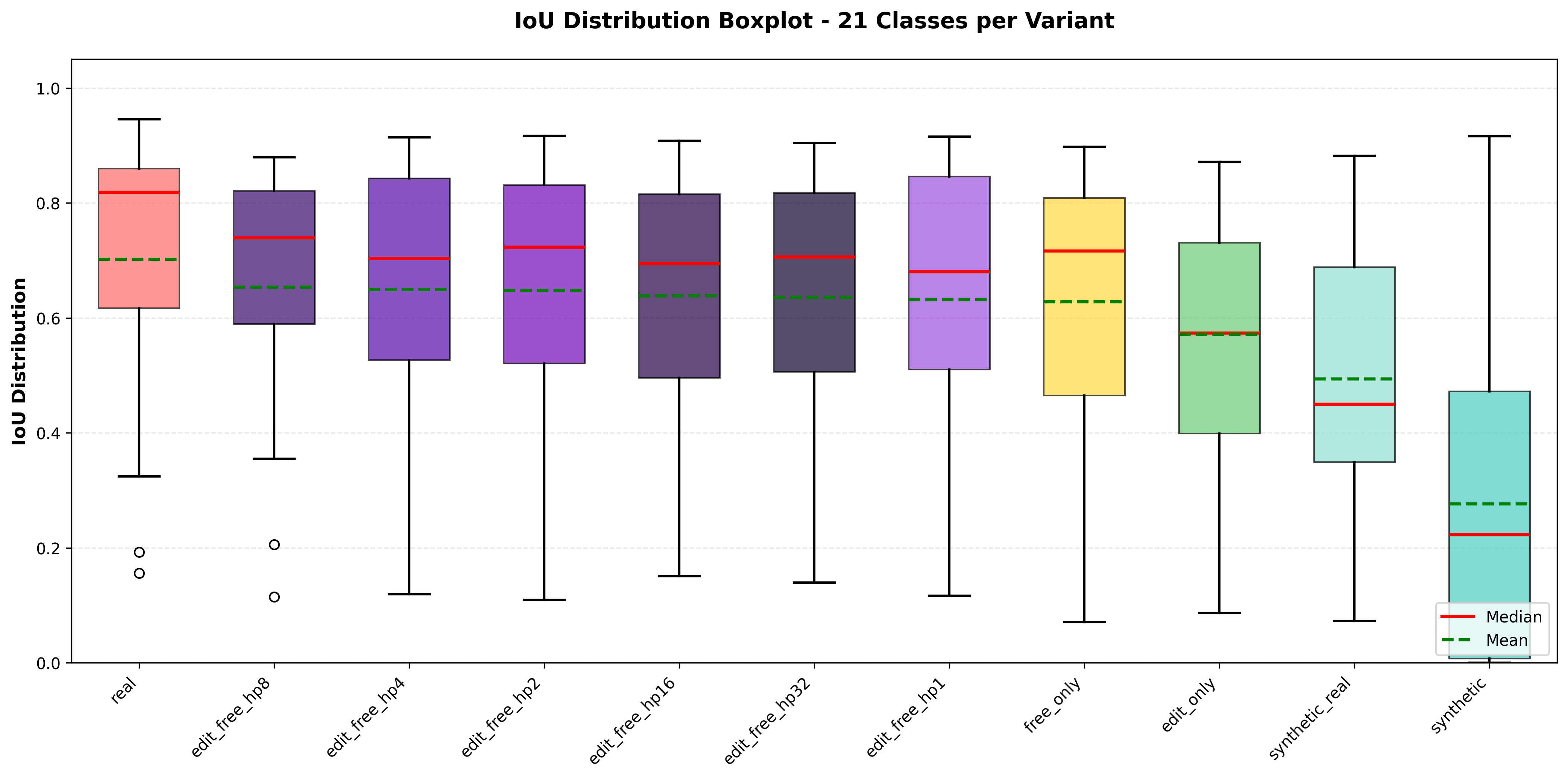}
\caption{Per-class IoU distribution across experiments (boxes show middle 50\% of classes, whiskers 1.5 IQR, dots are outliers).}
\label{fig:perclass_box}
\end{figure}

\begin{figure}[H]
\centering
\includegraphics[width=\linewidth]{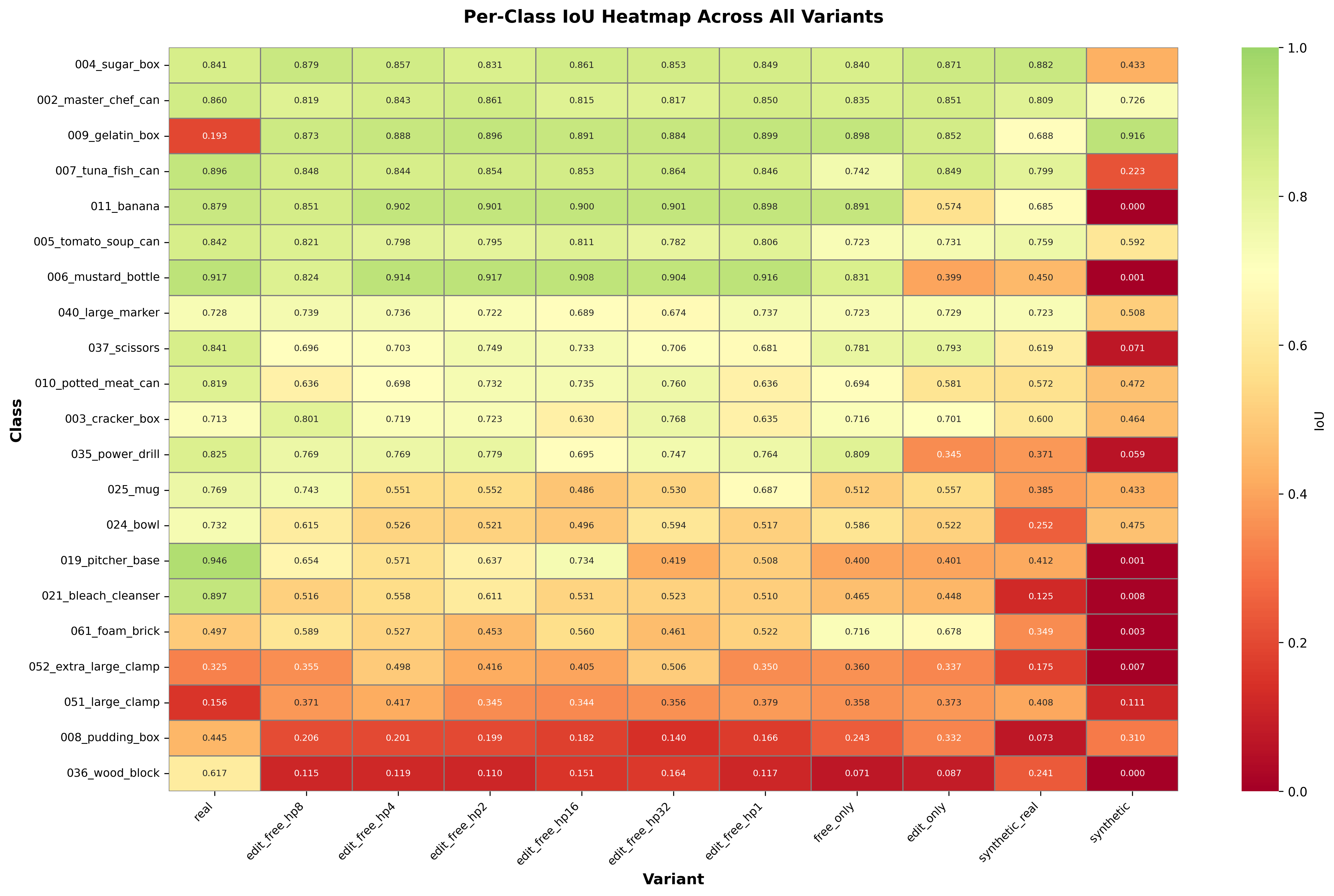}
\caption{Per-class IoU heatmap on YCB-V across experiments.}
\label{fig:heatmap}
\end{figure}

\FloatBarrier
\section{Notation}\label{app:notation}

\begin{table}[!h]
\caption{Notation used in the paper.}\label{tab:notation}
\centering
\small
\setlength{\tabcolsep}{3pt}
\begin{tabular}{>{\raggedright\arraybackslash}p{0.15\linewidth} >{\raggedright\arraybackslash}p{0.45\linewidth} >{\raggedright\arraybackslash}p{0.30\linewidth}}
\toprule
Symbol & Meaning & Notes \\
\midrule
$\mathcal{S}\,(p_S)$ & Synthetic domain (distribution) & \\
$\mathcal{R}\,(p_R)$ & Real domain (distribution) & \\
$x_s$ & input synthetic image $x_s$ & Source image for Syn2Real translation \\
$x_R$ & unpaired real image $x_R$ & Unpaired real reference \\
$y_R\,(y)$ & final output $y$ & Output of Syn2Real translation \\
$y_{\text{edit}}$ & editing-branch output $y_{\text{edit}}$ & Low-frequency base \\
$y_{\text{free}}$ & free-branch output $y_{\text{free}}$ & Residual generator output \\
$y_L, y_H$ & Low/high-frequency components & Relative frequency bands \\
$LP(\cdot)$ & Low-pass operator & \\
$\mathcal{L}$ & Loss scalar & Loss terms; see Sec.~\ref{sec:method-loss} \\
\bottomrule
\end{tabular}
\end{table}
\FloatBarrier

\end{appendices}

\clearpage
\FloatBarrier
\bibliography{sn-bibliography}


\begin{thebibliography}{22}
\ifx \bisbn   \undefined \def \bisbn  #1{ISBN #1}\fi
\ifx \binits  \undefined \def \binits#1{#1}\fi
\ifx \bauthor  \undefined \def \bauthor#1{#1}\fi
\ifx \batitle  \undefined \def \batitle#1{#1}\fi
\ifx \bjtitle  \undefined \def \bjtitle#1{#1}\fi
\ifx \bvolume  \undefined \def \bvolume#1{\textbf{#1}}\fi
\ifx \byear  \undefined \def \byear#1{#1}\fi
\ifx \bissue  \undefined \def \bissue#1{#1}\fi
\ifx \bfpage  \undefined \def \bfpage#1{#1}\fi
\ifx \blpage  \undefined \def \blpage #1{#1}\fi
\ifx \burl  \undefined \def \burl#1{\textsf{#1}}\fi
\ifx \doiurl  \undefined \def \doiurl#1{\url{https://doi.org/#1}}\fi
\ifx \betal  \undefined \def \betal{\textit{et al.}}\fi
\ifx \binstitute  \undefined \def \binstitute#1{#1}\fi
\ifx \binstitutionaled  \undefined \def \binstitutionaled#1{#1}\fi
\ifx \bctitle  \undefined \def \bctitle#1{#1}\fi
\ifx \beditor  \undefined \def \beditor#1{#1}\fi
\ifx \bpublisher  \undefined \def \bpublisher#1{#1}\fi
\ifx \bbtitle  \undefined \def \bbtitle#1{#1}\fi
\ifx \bedition  \undefined \def \bedition#1{#1}\fi
\ifx \bseriesno  \undefined \def \bseriesno#1{#1}\fi
\ifx \blocation  \undefined \def \blocation#1{#1}\fi
\ifx \bsertitle  \undefined \def \bsertitle#1{#1}\fi
\ifx \bsnm \undefined \def \bsnm#1{#1}\fi
\ifx \bsuffix \undefined \def \bsuffix#1{#1}\fi
\ifx \bparticle \undefined \def \bparticle#1{#1}\fi
\ifx \barticle \undefined \def \barticle#1{#1}\fi
\bibcommenthead
\ifx \bconfdate \undefined \def \bconfdate #1{#1}\fi
\ifx \botherref \undefined \def \botherref #1{#1}\fi
\ifx \url \undefined \def \url#1{\textsf{#1}}\fi
\ifx \bchapter \undefined \def \bchapter#1{#1}\fi
\ifx \bbook \undefined \def \bbook#1{#1}\fi
\ifx \bcomment \undefined \def \bcomment#1{#1}\fi
\ifx \oauthor \undefined \def \oauthor#1{#1}\fi
\ifx \citeauthoryear \undefined \def \citeauthoryear#1{#1}\fi
\ifx \endbibitem  \undefined \def \endbibitem {}\fi
\ifx \bconflocation  \undefined \def \bconflocation#1{#1}\fi
\ifx \arxivurl  \undefined \def \arxivurl#1{\textsf{#1}}\fi
\csname PreBibitemsHook\endcsname

\bibitem[\protect\citeauthoryear{Ren et~al.}{2015}]{ren2015fasterrcnn}
\begin{bchapter}
\bauthor{\bsnm{Ren}, \binits{S.}},
\bauthor{\bsnm{He}, \binits{K.}},
\bauthor{\bsnm{Girshick}, \binits{R.}},
\bauthor{\bsnm{Sun}, \binits{J.}}:
\bctitle{Faster {R-CNN}: Towards real-time object detection with region
  proposal networks}.
In: \bbtitle{Advances in Neural Information Processing Systems 28 ({NeurIPS})},
pp. \bfpage{91}--\blpage{99}
(\byear{2015}).
\burl{https://proceedings.neurips.cc/paper/2015/hash/14bfa6bb14875e45bba028a21ed38046-Abstract.html}
\end{bchapter}
\endbibitem

\bibitem[\protect\citeauthoryear{Chen et~al.}{2018}]{chen2018deeplabv3}
\begin{bchapter}
\bauthor{\bsnm{Chen}, \binits{L.-C.}},
\bauthor{\bsnm{Zhu}, \binits{Y.}},
\bauthor{\bsnm{Papandreou}, \binits{G.}},
\bauthor{\bsnm{Schroff}, \binits{F.}},
\bauthor{\bsnm{Adam}, \binits{H.}}:
\bctitle{Encoder-decoder with atrous separable convolution for semantic image
  segmentation}.
In: \bbtitle{Computer Vision -- {ECCV} 2018},
pp. \bfpage{833}--\blpage{851}
(\byear{2018}).
\doiurl{10.1007/978-3-030-01234-2_49}
\end{bchapter}
\endbibitem

\bibitem[\protect\citeauthoryear{Xiang et~al.}{2018}]{xiang2018posecnn}
\begin{bchapter}
\bauthor{\bsnm{Xiang}, \binits{Y.}},
\bauthor{\bsnm{Schmidt}, \binits{T.}},
\bauthor{\bsnm{Narayanan}, \binits{V.}},
\bauthor{\bsnm{Fox}, \binits{D.}}:
\bctitle{{PoseCNN}: A convolutional neural network for {6D} object pose
  estimation in cluttered scenes}.
In: \bbtitle{Proceedings of Robotics: Science and Systems ({RSS}) XIV}
(\byear{2018}).
\doiurl{10.15607/RSS.2018.XIV.019} .
\burl{https://www.roboticsproceedings.org/rss14/p19.html}
\end{bchapter}
\endbibitem

\bibitem[\protect\citeauthoryear{Guan et~al.}{2024}]{guan2024survey}
\begin{barticle}
\bauthor{\bsnm{Guan}, \binits{J.}},
\bauthor{\bsnm{Hao}, \binits{Y.}},
\bauthor{\bsnm{Wu}, \binits{Q.}},
\bauthor{\bsnm{Li}, \binits{S.}},
\bauthor{\bsnm{Fang}, \binits{Y.}}:
\batitle{A survey of {6DoF} object pose estimation methods for different
  application scenarios}.
\bjtitle{Sensors}
\bvolume{24}(\bissue{4}),
\bfpage{1076}
(\byear{2024})
\doiurl{10.3390/S24041076}
\end{barticle}
\endbibitem

\bibitem[\protect\citeauthoryear{Tobin et~al.}{2017}]{tobin2017domain}
\begin{bchapter}
\bauthor{\bsnm{Tobin}, \binits{J.}},
\bauthor{\bsnm{Fong}, \binits{R.}},
\bauthor{\bsnm{Ray}, \binits{A.}},
\bauthor{\bsnm{Schneider}, \binits{J.}},
\bauthor{\bsnm{Zaremba}, \binits{W.}},
\bauthor{\bsnm{Abbeel}, \binits{P.}}:
\bctitle{Domain randomization for transferring deep neural networks from
  simulation to the real world}.
In: \bbtitle{2017 {IEEE/RSJ} International Conference on Intelligent Robots and
  Systems ({IROS})},
pp. \bfpage{23}--\blpage{30}
(\byear{2017}).
\doiurl{10.1109/IROS.2017.8202133}
\end{bchapter}
\endbibitem

\bibitem[\protect\citeauthoryear{Bousmalis et~al.}{2017}]{bousmalis2017pixel}
\begin{bchapter}
\bauthor{\bsnm{Bousmalis}, \binits{K.}},
\bauthor{\bsnm{Silberman}, \binits{N.}},
\bauthor{\bsnm{Dohan}, \binits{D.}},
\bauthor{\bsnm{Erhan}, \binits{D.}},
\bauthor{\bsnm{Krishnan}, \binits{D.}}:
\bctitle{Unsupervised pixel-level domain adaptation with generative adversarial
  networks}.
In: \bbtitle{2017 {IEEE} Conference on Computer Vision and Pattern Recognition
  ({CVPR})},
pp. \bfpage{95}--\blpage{104}
(\byear{2017}).
\doiurl{10.1109/CVPR.2017.18}
\end{bchapter}
\endbibitem

\bibitem[\protect\citeauthoryear{Hoffman et~al.}{2018}]{hoffman2018cycada}
\begin{bchapter}
\bauthor{\bsnm{Hoffman}, \binits{J.}},
\bauthor{\bsnm{Tzeng}, \binits{E.}},
\bauthor{\bsnm{Park}, \binits{T.}},
\bauthor{\bsnm{Zhu}, \binits{J.-Y.}},
\bauthor{\bsnm{Isola}, \binits{P.}},
\bauthor{\bsnm{Saenko}, \binits{K.}},
\bauthor{\bsnm{Efros}, \binits{A.A.}},
\bauthor{\bsnm{Darrell}, \binits{T.}}:
\bctitle{{CyCADA}: Cycle-consistent adversarial domain adaptation}.
In: \bbtitle{Proceedings of the 35th International Conference on Machine
  Learning ({ICML})}.
\bsertitle{Proceedings of Machine Learning Research},
vol. \bseriesno{80},
pp. \bfpage{1989}--\blpage{1998}
(\byear{2018}).
\burl{https://proceedings.mlr.press/v80/hoffman18a.html}
\end{bchapter}
\endbibitem

\bibitem[\protect\citeauthoryear{Safayani et~al.}{2025}]{safayani2025review}
\begin{botherref}
\oauthor{\bsnm{Safayani}, \binits{M.}},
\oauthor{\bsnm{Mirzapour}, \binits{B.}},
\oauthor{\bsnm{Aghaebrahimiyan}, \binits{H.}},
\oauthor{\bsnm{Salehi}, \binits{N.}},
\oauthor{\bsnm{Ravaee}, \binits{H.}}:
Unpaired Image-to-Image Translation with Content Preserving Perspective: A
  Review.
arXiv preprint arXiv:2502.08667
(2025).
\doiurl{10.48550/arXiv.2502.08667} .
\url{https://arxiv.org/abs/2502.08667}
\end{botherref}
\endbibitem

\bibitem[\protect\citeauthoryear{Imbusch et~al.}{2022}]{imbusch2022cut}
\begin{bchapter}
\bauthor{\bsnm{Imbusch}, \binits{B.T.}},
\bauthor{\bsnm{Schwarz}, \binits{M.}},
\bauthor{\bsnm{Behnke}, \binits{S.}}:
\bctitle{Synthetic-to-real domain adaptation using contrastive unpaired
  translation}.
In: \bbtitle{2022 {IEEE} International Conference on Automation Science and
  Engineering ({CASE})},
pp. \bfpage{595}--\blpage{602}
(\byear{2022}).
\doiurl{10.1109/CASE49997.2022.9926640}
\end{bchapter}
\endbibitem

\bibitem[\protect\citeauthoryear{Guo et~al.}{2022}]{guo2022structure}
\begin{bchapter}
\bauthor{\bsnm{Guo}, \binits{J.}},
\bauthor{\bsnm{Li}, \binits{J.}},
\bauthor{\bsnm{Fu}, \binits{H.}},
\bauthor{\bsnm{Gong}, \binits{M.}},
\bauthor{\bsnm{Zhang}, \binits{K.}},
\bauthor{\bsnm{Tao}, \binits{D.}}:
\bctitle{Alleviating semantics distortion in unsupervised low-level
  image-to-image translation via structure consistency constraint}.
In: \bbtitle{2022 {IEEE/CVF} Conference on Computer Vision and Pattern
  Recognition ({CVPR})},
pp. \bfpage{18228}--\blpage{18238}
(\byear{2022}).
\doiurl{10.1109/CVPR52688.2022.01771}
\end{bchapter}
\endbibitem

\bibitem[\protect\citeauthoryear{Zhu et~al.}{2017}]{zhu2017cyclegan}
\begin{bchapter}
\bauthor{\bsnm{Zhu}, \binits{J.-Y.}},
\bauthor{\bsnm{Park}, \binits{T.}},
\bauthor{\bsnm{Isola}, \binits{P.}},
\bauthor{\bsnm{Efros}, \binits{A.A.}}:
\bctitle{Unpaired image-to-image translation using cycle-consistent adversarial
  networks}.
In: \bbtitle{2017 {IEEE} International Conference on Computer Vision ({ICCV})},
pp. \bfpage{2242}--\blpage{2251}
(\byear{2017}).
\doiurl{10.1109/ICCV.2017.244}
\end{bchapter}
\endbibitem

\bibitem[\protect\citeauthoryear{Park et~al.}{2020}]{park2020cut}
\begin{bchapter}
\bauthor{\bsnm{Park}, \binits{T.}},
\bauthor{\bsnm{Efros}, \binits{A.A.}},
\bauthor{\bsnm{Zhang}, \binits{R.}},
\bauthor{\bsnm{Zhu}, \binits{J.-Y.}}:
\bctitle{Contrastive learning for unpaired image-to-image translation}.
In: \bbtitle{Computer Vision -- {ECCV} 2020},
pp. \bfpage{319}--\blpage{345}
(\byear{2020}).
\doiurl{10.1007/978-3-030-58545-7_19}
\end{bchapter}
\endbibitem

\bibitem[\protect\citeauthoryear{Yang and Soatto}{2020}]{yang2020fda}
\begin{bchapter}
\bauthor{\bsnm{Yang}, \binits{Y.}},
\bauthor{\bsnm{Soatto}, \binits{S.}}:
\bctitle{{FDA}: Fourier domain adaptation for semantic segmentation}.
In: \bbtitle{2020 {IEEE/CVF} Conference on Computer Vision and Pattern
  Recognition ({CVPR})},
pp. \bfpage{4084}--\blpage{4094}
(\byear{2020}).
\doiurl{10.1109/CVPR42600.2020.00414}
\end{bchapter}
\endbibitem

\bibitem[\protect\citeauthoryear{Liang et~al.}{2021}]{liang2021lptn}
\begin{bchapter}
\bauthor{\bsnm{Liang}, \binits{J.}},
\bauthor{\bsnm{Zeng}, \binits{H.}},
\bauthor{\bsnm{Zhang}, \binits{L.}}:
\bctitle{High-resolution photorealistic image translation in real-time: A
  laplacian pyramid translation network}.
In: \bbtitle{2021 {IEEE/CVF} Conference on Computer Vision and Pattern
  Recognition ({CVPR})},
pp. \bfpage{9387}--\blpage{9395}
(\byear{2021}).
\doiurl{10.1109/CVPR46437.2021.00927}
\end{bchapter}
\endbibitem

\bibitem[\protect\citeauthoryear{Cai et~al.}{2021}]{cai2021frequency}
\begin{bchapter}
\bauthor{\bsnm{Cai}, \binits{M.}},
\bauthor{\bsnm{Zhang}, \binits{H.}},
\bauthor{\bsnm{Huang}, \binits{H.}},
\bauthor{\bsnm{Geng}, \binits{Q.}},
\bauthor{\bsnm{Li}, \binits{Y.}},
\bauthor{\bsnm{Huang}, \binits{G.}}:
\bctitle{Frequency domain image translation: More photo-realistic, better
  identity-preserving}.
In: \bbtitle{2021 {IEEE/CVF} International Conference on Computer Vision
  ({ICCV})},
pp. \bfpage{13910}--\blpage{13920}
(\byear{2021}).
\doiurl{10.1109/ICCV48922.2021.01367}
\end{bchapter}
\endbibitem

\bibitem[\protect\citeauthoryear{Rahaman et~al.}{2019}]{rahaman2019spectral}
\begin{bchapter}
\bauthor{\bsnm{Rahaman}, \binits{N.}},
\bauthor{\bsnm{Baratin}, \binits{A.}},
\bauthor{\bsnm{Arpit}, \binits{D.}},
\bauthor{\bsnm{Draxler}, \binits{F.}},
\bauthor{\bsnm{Lin}, \binits{M.}},
\bauthor{\bsnm{Hamprecht}, \binits{F.A.}},
\bauthor{\bsnm{Bengio}, \binits{Y.}},
\bauthor{\bsnm{Courville}, \binits{A.}}:
\bctitle{On the spectral bias of neural networks}.
In: \bbtitle{Proceedings of the 36th International Conference on Machine
  Learning ({ICML})}.
\bsertitle{Proceedings of Machine Learning Research},
vol. \bseriesno{97},
pp. \bfpage{5301}--\blpage{5310}
(\byear{2019}).
\burl{https://proceedings.mlr.press/v97/rahaman19a.html}
\end{bchapter}
\endbibitem

\bibitem[\protect\citeauthoryear{Jiang et~al.}{2021}]{jiang2021ffl}
\begin{bchapter}
\bauthor{\bsnm{Jiang}, \binits{L.}},
\bauthor{\bsnm{Dai}, \binits{B.}},
\bauthor{\bsnm{Wu}, \binits{W.}},
\bauthor{\bsnm{Loy}, \binits{C.C.}}:
\bctitle{Focal frequency loss for image reconstruction and synthesis}.
In: \bbtitle{2021 {IEEE/CVF} International Conference on Computer Vision
  ({ICCV})},
pp. \bfpage{13899}--\blpage{13909}
(\byear{2021}).
\doiurl{10.1109/ICCV48922.2021.01366}
\end{bchapter}
\endbibitem

\bibitem[\protect\citeauthoryear{Isola et~al.}{2017}]{isola2017pix2pix}
\begin{bchapter}
\bauthor{\bsnm{Isola}, \binits{P.}},
\bauthor{\bsnm{Zhu}, \binits{J.-Y.}},
\bauthor{\bsnm{Zhou}, \binits{T.}},
\bauthor{\bsnm{Efros}, \binits{A.A.}}:
\bctitle{Image-to-image translation with conditional adversarial networks}.
In: \bbtitle{2017 {IEEE} Conference on Computer Vision and Pattern Recognition
  ({CVPR})},
pp. \bfpage{5967}--\blpage{5976}
(\byear{2017}).
\doiurl{10.1109/CVPR.2017.632}
\end{bchapter}
\endbibitem

\bibitem[\protect\citeauthoryear{Miyato et~al.}{2018}]{miyato2018spectral}
\begin{bchapter}
\bauthor{\bsnm{Miyato}, \binits{T.}},
\bauthor{\bsnm{Kataoka}, \binits{T.}},
\bauthor{\bsnm{Koyama}, \binits{M.}},
\bauthor{\bsnm{Yoshida}, \binits{Y.}}:
\bctitle{Spectral normalization for generative adversarial networks}.
In: \bbtitle{International Conference on Learning Representations ({ICLR})}
(\byear{2018}).
\burl{https://openreview.net/forum?id=B1QRgziT-}
\end{bchapter}
\endbibitem

\bibitem[\protect\citeauthoryear{Goodfellow et~al.}{2014}]{goodfellow2014gan}
\begin{bchapter}
\bauthor{\bsnm{Goodfellow}, \binits{I.J.}},
\bauthor{\bsnm{Pouget-Abadie}, \binits{J.}},
\bauthor{\bsnm{Mirza}, \binits{M.}},
\bauthor{\bsnm{Xu}, \binits{B.}},
\bauthor{\bsnm{Warde-Farley}, \binits{D.}},
\bauthor{\bsnm{Ozair}, \binits{S.}},
\bauthor{\bsnm{Courville}, \binits{A.C.}},
\bauthor{\bsnm{Bengio}, \binits{Y.}}:
\bctitle{Generative adversarial nets}.
In: \bbtitle{Advances in Neural Information Processing Systems 27 ({NeurIPS})},
pp. \bfpage{2672}--\blpage{2680}
(\byear{2014}).
\burl{https://proceedings.neurips.cc/paper/5423-generative-adversarial-nets}
\end{bchapter}
\endbibitem

\bibitem[\protect\citeauthoryear{Jocher et~al.}{2026}]{jocher2026yolo}
\begin{botherref}
\oauthor{\bsnm{Jocher}, \binits{G.}},
\oauthor{\bsnm{Qiu}, \binits{J.}},
\oauthor{\bsnm{Chaurasia}, \binits{A.}}:
Ultralytics {YOLO}.
Zenodo.
Version v8.3.248
(2026).
\doiurl{10.5281/zenodo.18147344} .
\url{https://doi.org/10.5281/zenodo.18147344}
\end{botherref}
\endbibitem

\bibitem[\protect\citeauthoryear{Hodan et~al.}{2018}]{hodan2018bop}
\begin{bchapter}
\bauthor{\bsnm{Hodan}, \binits{T.}},
\bauthor{\bsnm{Michel}, \binits{F.}},
\bauthor{\bsnm{Brachmann}, \binits{E.}},
\bauthor{\bsnm{Kehl}, \binits{W.}},
\bauthor{\bsnm{Buch}, \binits{A.G.}},
\bauthor{\bsnm{Kraft}, \binits{D.}},
\bauthor{\bsnm{Drost}, \binits{B.}},
\bauthor{\bsnm{Vidal}, \binits{J.}},
\bauthor{\bsnm{Ihrke}, \binits{S.}},
\bauthor{\bsnm{Zabulis}, \binits{X.}},
\bauthor{\bsnm{Sahin}, \binits{C.}},
\bauthor{\bsnm{Manhardt}, \binits{F.}},
\bauthor{\bsnm{Tombari}, \binits{F.}},
\bauthor{\bsnm{Kim}, \binits{T.-K.}},
\bauthor{\bsnm{Matas}, \binits{J.}},
\bauthor{\bsnm{Rother}, \binits{C.}}:
\bctitle{{BOP}: Benchmark for {6D} object pose estimation}.
In: \bbtitle{Computer Vision -- {ECCV} 2018},
pp. \bfpage{19}--\blpage{35}
(\byear{2018}).
\doiurl{10.1007/978-3-030-01249-6_2}
\end{bchapter}
\endbibitem

\end{thebibliography}

\end{document}